\documentclass{article}
\usepackage{ijcai23}
\usepackage{times}
\usepackage{soul}
\usepackage{url}
\usepackage[hidelinks]{hyperref}
\usepackage[utf8]{inputenc}
\usepackage{caption}
\usepackage{graphicx}
\usepackage{amsmath}
\usepackage{amsthm}
\usepackage{booktabs}
\usepackage{algorithm}
\usepackage{algorithmic}
\usepackage[switch]{lineno}

\usepackage{xcolor}
\usepackage{makecell}
\usepackage{amsmath,amssymb}
\usepackage{tabularx}
\usepackage{array, makecell}
\usepackage{tablefootnote}
\usepackage{caption}
\usepackage{subcaption}
\usepackage[font=small]{caption}
\usepackage{pifont}
\usepackage[english]{babel}
\newcommand{\cmark}{\ding{51}}%
\newcommand{\xmark}{\ding{55}}%
\DeclareMathOperator{\E}{\mathbb{E}}

\title{Answer Mining from a Pool of Images: Towards \\Retrieval-Based Visual Question Answering}
\author{
Abhirama Subramanyam Penamakuri$^1$
\and
Manish Gupta$^2$\and
Mithun Das Gupta$^{2}$\and
Anand Mishra$^1$
\affiliations
$^1$Indian Institute of Technology Jodhpur\\
$^2$Microsoft
\emails
\{penamakuri.1, mishra\}@iitj.ac.in,
\{gmanish,migupta\}@microsoft.com
}

\begin{document}

\maketitle

\begin{abstract}
We study visual question answering in a setting where the answer has to be mined from a pool of relevant and irrelevant images given as a context. For such a setting, a model must first retrieve relevant images from the pool and answer the question from these retrieved images. We refer to this problem as retrieval-based visual question answering (or \textsc{RetVQA} in short). The \textsc{RetVQA} is distinctively different and more challenging than the traditionally-studied Visual Question Answering (VQA), where a given question has to be answered with a single relevant image in context. Towards solving the \textsc{RetVQA} task, we propose a unified \underline{M}ulti \underline{I}mage \underline{BART} (MI-BART) that takes a question and retrieved images using our relevance encoder for free-form fluent answer generation. Further, we introduce the largest dataset in this space, namely \textsc{RetVQA}, which has the following salient features: multi-image and retrieval requirement for VQA, metadata-independent questions over a pool of heterogeneous images, expecting a mix of classification-oriented and open-ended generative answers. Our proposed framework achieves an accuracy of 76.5\% and a fluency of 79.3\% on the proposed dataset, namely \textsc{RetVQA} and also outperforms state-of-the-art methods by 4.9\% and 11.8\% on the image segment of the publicly available WebQA dataset on the accuracy and fluency metrics, respectively.
\end{abstract}
\section{Introduction}
\begin{figure}[t!]
\centering
  \includegraphics[width=0.8\columnwidth]{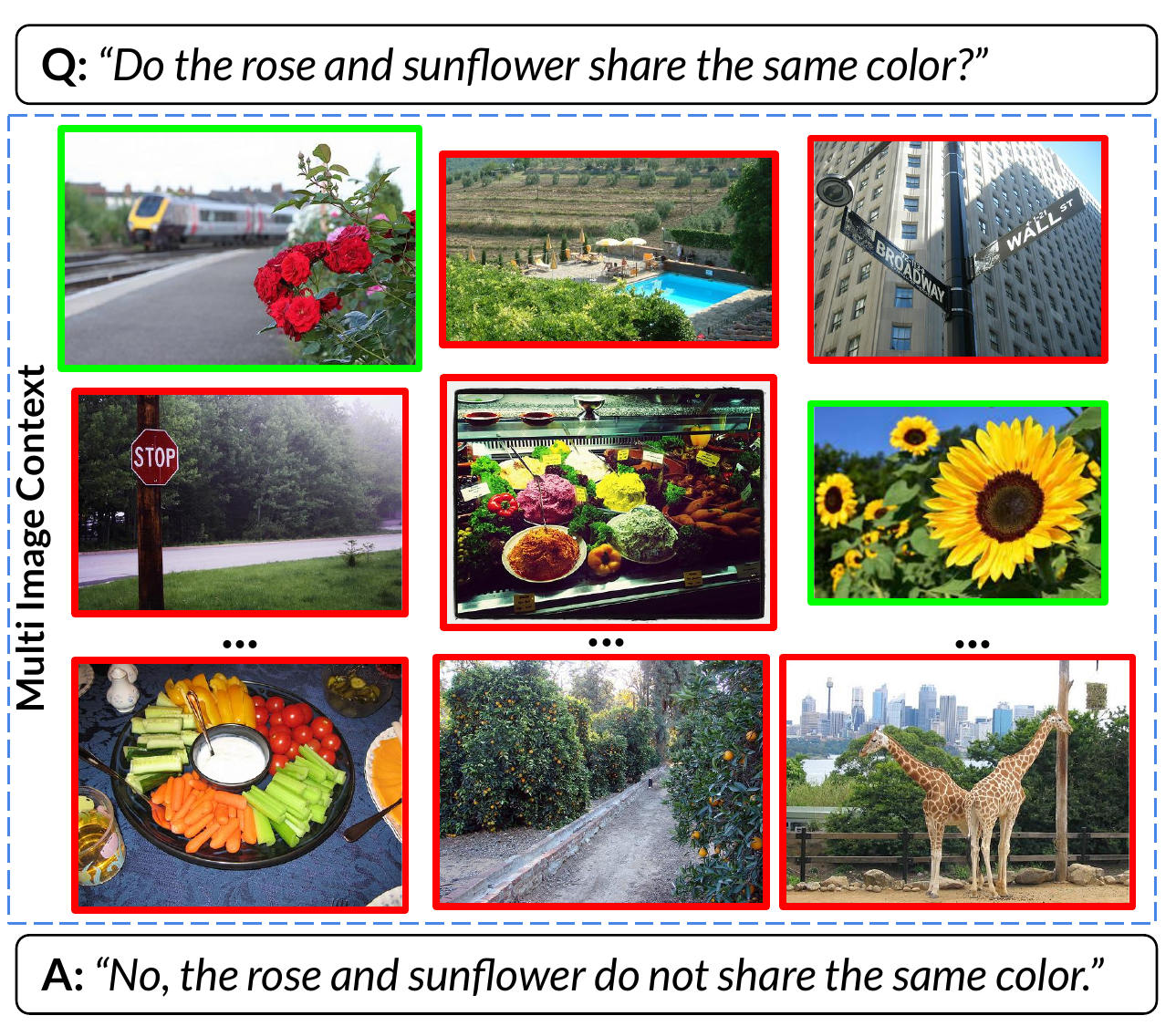}
  \caption{Given a question and a pool of images (multi-image context), \textsc{RetVQA} task involves two stages: (i) retrieve the relevant images from the pool, and (ii) generate a free-form natural language answer by reasoning over the retrieved relevant images as context.}
  \label{fig:goal}
\end{figure}

Question Answering (QA) over textual as well as visual data has been an active area of research~\cite{retQA,guo2021bilinear}. In text-based QA, the research focus has recently shifted from highly-explored QA on a single paragraph such as SQuAD~\cite{squad} to a setting where mining answers from a huge corpus of documents is a requirement~\cite{ahmad2019reqa,retQA}. On the contrary, visual question answering (VQA)~\cite{antol2015vqa} literature has so far largely restricted itself to answering questions about a given relevant visual context (often a single image). However, this does not necessarily suffice to satisfy our information needs since the information may be spread across multiple images and may not be present in some images. For example, consider a natural language question `\textit{Do the rose and sunflower share the same color?}', answering such a question from a pool of images as visual context (refer Figure~\ref{fig:goal}), requires a model to first retrieve relevant images and then perform visio-lingual reasoning on the retrieved images to arrive at a fluent free-form natural language answer. We refer to this problem as \textsc{RetVQA} or retrieval-based visual question answering. The \textsc{RetVQA} setting has potential applications in question answering on web images, e-commerce, environmental monitoring, and health care, among others, e.g., multiple images of a particular area can be analyzed to monitor environmental changes over time; multiple MRI or CT scans of a patient's brain need to be analyzed to detect abnormalities, such as tumors.

For~\textsc{RetVQA}, the input is a pool of images with only a few images being relevant to the question. Close to our setting, there is some exciting progress in the recent literature~\cite{talmor2021multimodalqa,imageset_vqa,singh2021mimoqa,webQA21}. However, these works assume one or more of the following constraints: ``without requiring explicit retrieval", ``having classification-type fixed-vocabulary answers", ``assuming the availability of meta-data like \textit{WikiEntities, captions}", ``having a homogeneous yet limited number of images in the pool", and ``having only a small set of questions that need multiple images". Such constraints in the existing datasets point towards a need for a large-scale benchmark to study \textsc{RetVQA}. To this end, we present a \textit{derived} dataset prepared from Visual Genome~\cite{krishna2017visual}, leveraging its questions and annotations of images. We curate questions under different categories: (i) common attributes such as color, shape, and count, (ii) other object-attributes that include non-common attributes, e.g. length, material, and (iii) subject-object relationships, e.g., `eats', `left of'. Further, to facilitate benchmarking capabilities of the VQA models over open-ended answers, we curate questions under binary (yes/no) and open-ended answer categories. Note that the answers are free-form fluent in both the answer categories, e.g. `\textit{No, rose and sunflower do not share the same color}' (a binary answer); `\textit{The color of rose and sunflower is red and yellow, respectively}' (an open-ended generative answer). \textsc{RetVQA} dataset statistics and distribution across the question-answer categories are shown in Table~\ref{tab:moredatasetStats} and Table~\ref{tab:datasetStats}, respectively.

\begin{table}[!t]
\centering
\resizebox{0.7\columnwidth}{!}{

\begin{tabular}{ll}
\hline
\textbf{Measurement}&\textbf{Value}\\ 
\hline
\#Distinct questions & 418K \\
\#Distinct precise answers & 16,205 \\
Train set questions & 334K (80\%) \\
Val set questions & 41K (10\%) \\
Test set questions & 41K (10\%) \\ 
Avg question length (words) & 8.7 \\
Avg answer length (words) & 8.5\\ 
\#Distinct words in questions & 10,868 \\ 
\#Distinct words in answers & 9,278 \\ 
\#Avg relevant images per question & 2 \\
\#Avg irrelevant images per question & 24.5\\
\hline
\end{tabular}}
\caption{Key statistics for \textsc{RetVQA} dataset.}
\label{tab:moredatasetStats}
\end{table} 

Further, to solve the \textsc{RetVQA} task, a model must first retrieve the relevant images for the question and then consume the retrieved images as the context to answer the question. Towards this end, we present a unified \underline{M}ulti \underline{I}mage \underline{BART} that takes in the question along with the multi-image context retrieved using a relevance encoder to generate the free-form fluent natural-language answer. Our proposed framework, MI-BART, allows joint reasoning over multiple retrieved images along with the question to capture better semantics.

To summarize, our contributions are as follows: (i) We present \textsc{RetVQA}, a 20$\times$ larger dataset than the closest dataset~\cite{webQA21} in this setting. \textsc{RetVQA} dataset is prepared by leveraging questions and image annotations from Visual Genome. It emphasizes on multi-image, metadata-independent questions over a pool of heterogenous collections of images, expecting a mix of classification-oriented and generative answers. We strongly believe that the proposed task, curated dataset and benchmarks presented in this paper will pave the way for further research.
(ii) We present \underline{M}ulti \underline{I}mage \underline{BART} (MI-BART) - a unified method that reasons jointly over the retrieved multi-image context along with the question to generate a free-form fluent answer for the question. (iii) We perform extensive experiments to evaluate the performance of our proposed framework on \textsc{RetVQA} and the image segment of WebQA. Our approach clearly outperforms baseline approaches on \textsc{RetVQA} dataset and achieves state-of-the-art performance on the image segment of WebQA. We make our data and implementation publicly available.\footnote{\hyperlink{https://vl2g.github.io/projects/retvqa/}{https://vl2g.github.io/projects/retvqa/}}

\section{Related Work}
\label{sec:relWorks}
\noindent\paragraph{Visual and Multi-modal QA.}
Visual Question Answering (VQA) aims at answering a natural language question in the context of a relevant image~\cite{antol2015vqa}. This area has seen significant progress partly due to the introduction of several challenging datasets~\cite{malinowski2015ask,ren2015exploring,zhu2016visual7w,antol2015vqa,goyal2017making,johnson2017clevr,krishna2017visual}. Most methods for VQA either use a multimodal fusion of language and image embeddings~\cite{ren2015exploring,gao2015you,noh2016image,kembhavi2017you}, attention-based multimodal fusion~\cite{yang2016stacked,fukui2016multimodal,shih2016look,lu2016hierarchical,xiong2016dynamic} or neural module networks~\cite{andreas2016neural,hu2017learning}. More recently, knowledge-based VQA~\cite{shah2019kvqa,marino2019ok} has gained attention where external knowledge is used for answering visual questions. Contrary to these exciting works in VQA literature, our problem setting is distinctively different as we need to mine the answer from a collection of relevant as well as irrelevant images.

\begin{table}[!t]
\centering
\resizebox{0.65\columnwidth}{!}{
\begin{tabular}{lcc|c}
\hline
\textbf{Question Category}&{\textbf{Binary}}&\textbf{Open-ended}&\textbf{Total} \\ 
\hline
Color & 50K & 50K & 100K\\
Shape & 49K & 50K & 99K \\
Count & 50K & 50K & 100K\\ 
Object-attributes & 80K & - & 80K\\ 
Relation-based & - & 38K & 38K \\ 
\hline
Total & 229K & 188K & 418K \\
\hline
\end{tabular}}
\caption{Distribution of questions by various categories in \textsc{RetVQA} dataset. The answers are of two types: binary-generative and open-ended generative.}
\label{tab:datasetStats}
\end{table} 

\begin{table*}[!t]
\centering
\resizebox{0.9\textwidth}{!}{
\begin{tabular}{lccccccc}
\hline
\textbf{Dataset} & 
\textbf{\#Questions} & 
\makecell{\textbf{Retrieval}\\ \textbf{required}} & \makecell{\textbf{Heterogenous}\\ \textbf{images}} & \makecell{\textbf{Multi-image} \\ \textbf{reasoning}} & \makecell{\textbf{No Meta-data} \\ \textbf{assumption}} & 
\makecell{\textbf{Answer type}} & 
\makecell{\textbf{\% of questions that} \\ \textbf{need multiple images}} \\
\hline
MultimodalQA~\cite{talmor2021multimodalqa} & 2K & \xmark & \xmark & \cmark & \xmark~(WikiEntities) & Classification & 6.1\%\\
ISVQA~\cite{imageset_vqa} & 141K & \xmark & \xmark & \cmark & \cmark & Classification & 33\% \\
WebQA~\cite{webQA21} & 18K & \cmark & \cmark & \cmark & \xmark~(Captions) & Open-ended & 44\% \\
\textsc{RetVQA} (Ours) & \textbf{327K} & \cmark & \cmark & \cmark & \cmark & Open-ended & \textbf{100\%} \\
\hline
\end{tabular}}
\caption{Comparison of our curated dataset \textsc{RetVQA} with other relevant QA datasets. For Multimodal QA and WebQA datasets, we have considered their image-only modality questions subset.}
\label{tab:datasetComp}
\end{table*} 

Sharing a similar motivation as ours, the following tasks and accompanying datasets have been recently introduced in the literature: (i) MultimodalQA~\cite{talmor2021multimodalqa}, (ii) ISVQA~\cite{imageset_vqa}, and (iii) WebQA~\cite{webQA21}.
In MultimodalQA~\cite{talmor2021multimodalqa}, only a small part of the dataset (ImageListQ) is relevant to our setting; however, even on this subset, MultimodalQA assumes the availability of extra image metadata, i.e., table or WikiEntity linkage. Similarly, in ISVQA~\cite{imageset_vqa}, every question has a small set of homogeneous images as context. Since images are homogeneous, there is no need for explicit retrieval. Recently, WebQA~\cite{webQA21} dataset has been proposed to target such practical QA scenarios, where a question has to be answered in the context of multimodal sources; however, the images in the dataset have associated captions. All of these settings have differences from \textsc{RetVQA} in either usage of additional context, constraints on images in the collection, answer schema, or classification instead of generation. In another recent work MIMOQA~\cite{singh2021mimoqa}, only extractive question answering is performed. 
In terms of reasoning on more than one image, another related work is NLVR~\cite{suhr2019corpus}. However, it does not involve any retrieval and open-ended answer generation. Further, in terms of QA over multiple document images or video frames, related works are DocVQA~\cite{tito2021document} and VideoQA~\cite{lei2018tvqa,lei2020tvqa+,tapaswi2016movieqa}. DocVQA focuses on text-heavy document images, limiting visual reasoning, whereas the VideoQA task does not involve explicit retrieval and open-ended answer generation. 
Contrary to these works, our newly curated dataset is significantly larger, has no assumption of meta-data availability, and requires retrieval and reasoning over multiple images to arrive at an answer. 

\noindent\paragraph{Multi-modal modeling.}
Recently, multi-modal transformers such as VisualBERT~\cite{visualbert}, 
VilBERT~\cite{vilbert}, VILT~\cite{vilt}, LXMERT~\cite{tan2019lxmert}, OSCAR ~\cite{oscar}, UNITER \cite{chen2020uniter} have shown strong results on the downstream vision and language tasks. However, these encoder-based models are more suited for classification-style VQA settings. Multimodal transformers like VLP~\cite{VLP} and VLBart~\cite{vlbart} are pre-trained with sequence-to-sequence objectives and hence are more suitable for the current setting that requires the model to generate free-form natural language answers. We follow a similar approach by devising an encoder-decoder framework to jointly reason over multiple images along with the question.

\section{\textsc{RetVQA} Dataset}
\label{sec:dataset}

\begin{figure}
  \scriptsize
   \centering
   \begin{subfigure}[b]{0.45\columnwidth}
        \centering  \includegraphics[width=\columnwidth]{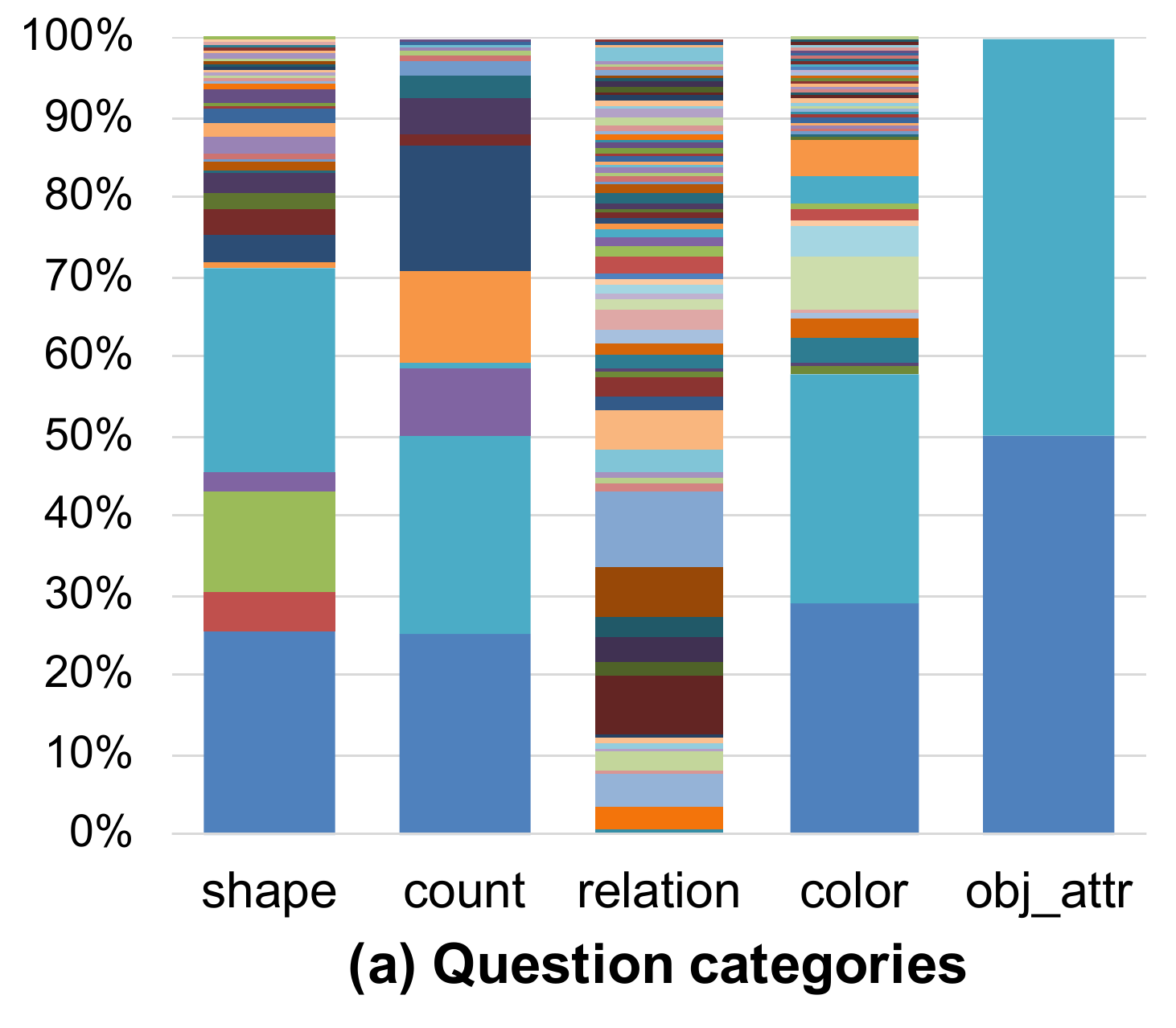}
     \end{subfigure}
     \begin{subfigure}[b]{0.45\columnwidth}
     \centering
     \includegraphics[width=\columnwidth]{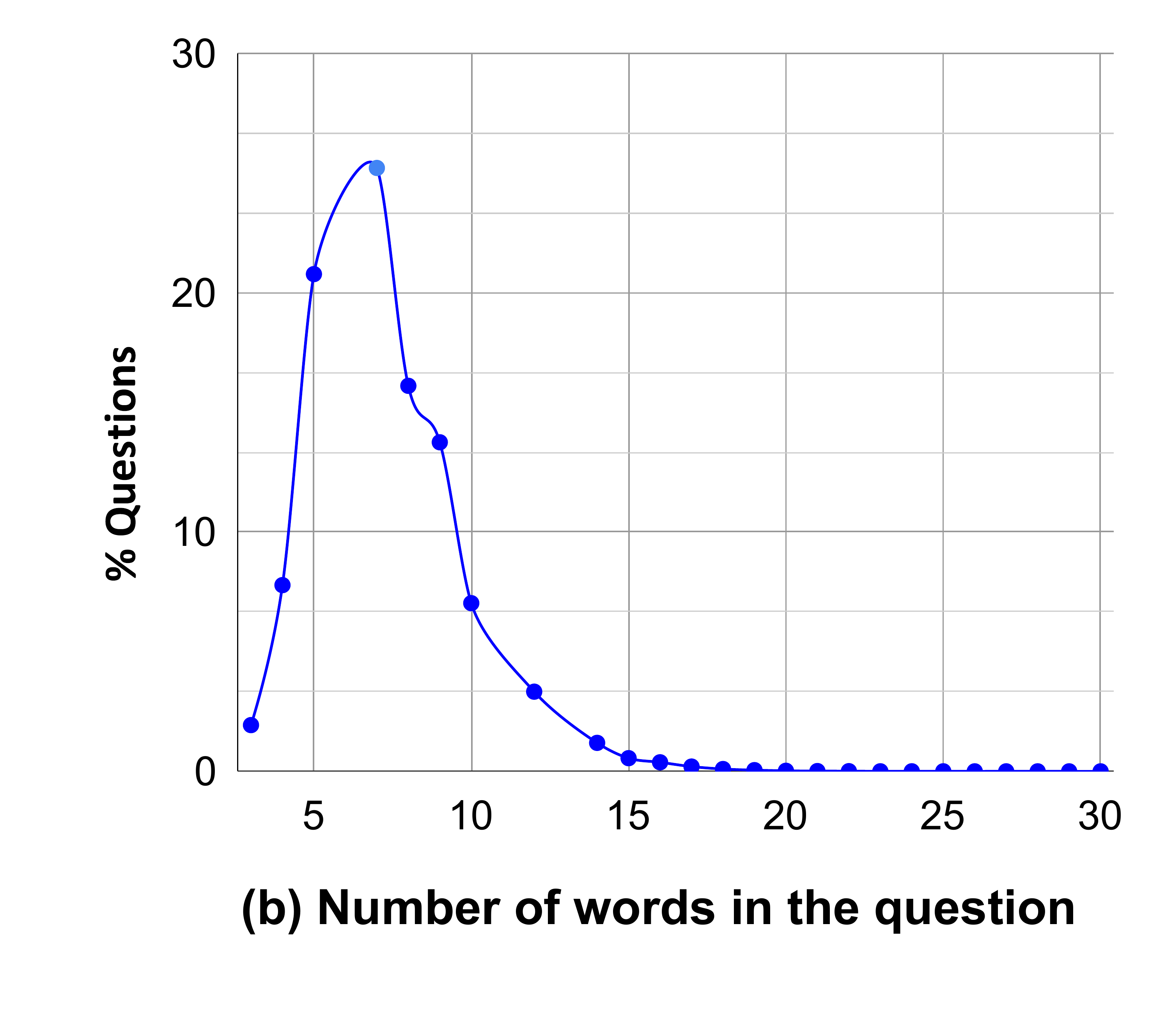}
     \end{subfigure}
     \caption{\textsc{RetVQA} questions and answers analysis: (a) Answer distribution over various question categories, (b) Distribution of the number of words across questions. 
     }
     \label{fig:datasetAnalysis}
\end{figure}

Traditionally, VQA datasets~\cite{antol2015vqa,goyal2017making,singh2019towards,talmor2021multimodalqa} assume that the context provided is always relevant to the question. 
\begin{figure}[t!]
\centering
  \includegraphics[width=0.65\columnwidth]{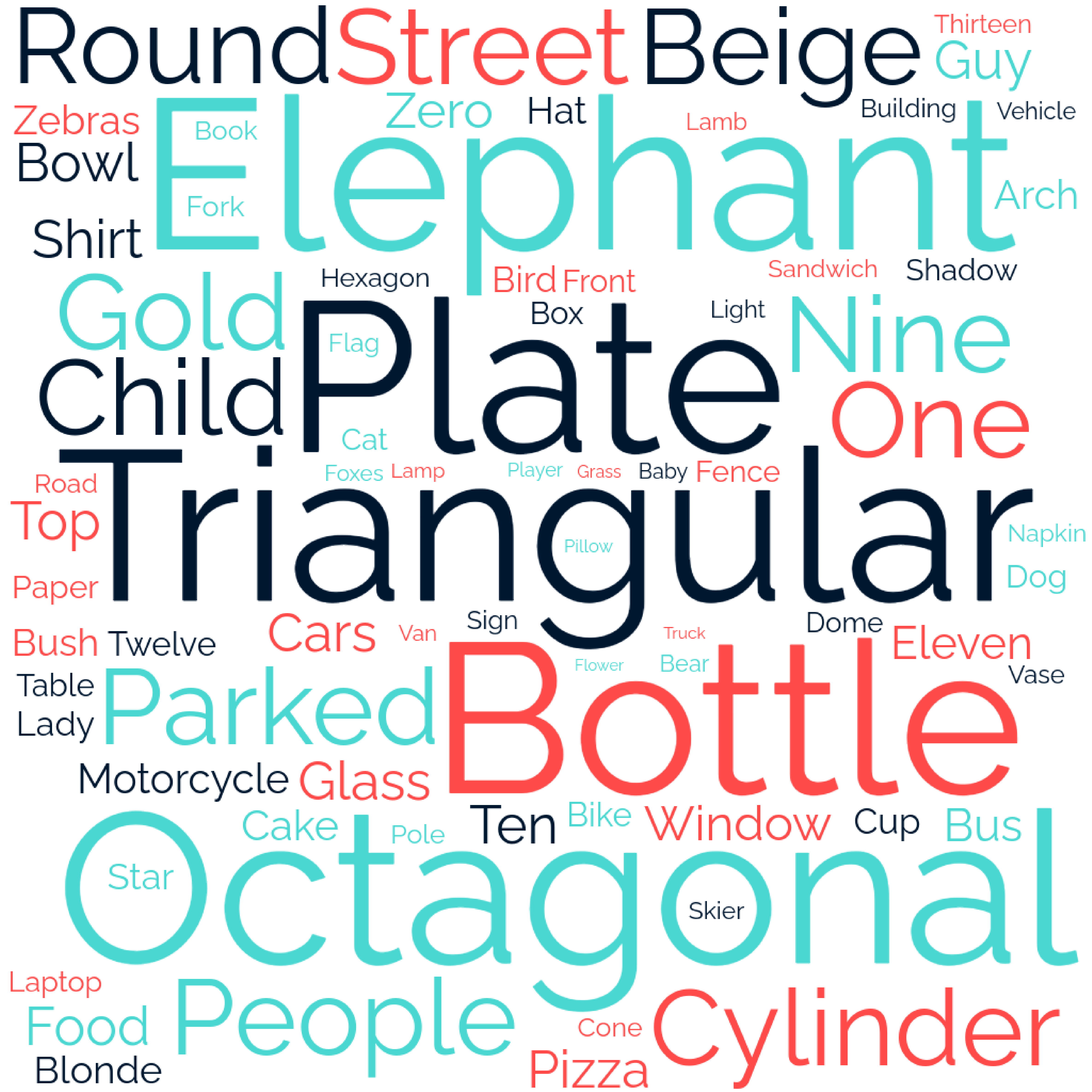}
  \caption{Word cloud of Top-80 frequent answers.}
  \label{fig:word_cloud}
\end{figure}
Recently,~\cite{webQA21} proposed a benchmark where given a question and a pool of multimodal sources (containing both image and text snippets as context), only a few of these sources are relevant to the question. Similar to our problem setup, it requires first retrieving the relevant context and then using it to answer the question. However, after carefully observing the WebQA dataset, we found that most questions include rare entities like `Maracana Stadium', `Minnetonka Rhododendron', etc. Such questions make the retrieval task of the problem non-trivial without auxiliary information about these images. Methods proposed in~\cite{webQA21} leverage metadata like image captions, which contain information like the name of the entity in the image. Further, a rule-based retrieval using word overlap of the question with the image caption for the retrieval task has an F1 score of 37, asserting our claim that retrieval is over-dependent on image metadata and not significantly on visual data. Further, the image-based subset of WebQA has a majority of samples (55.6\%) with one relevant image per question, thereby, a single image VQA method may perform equally well given the relevant image is retrieved.

\begin{figure*}[t!]
   \centering
   \scriptsize
  \includegraphics[width=0.97\textwidth]{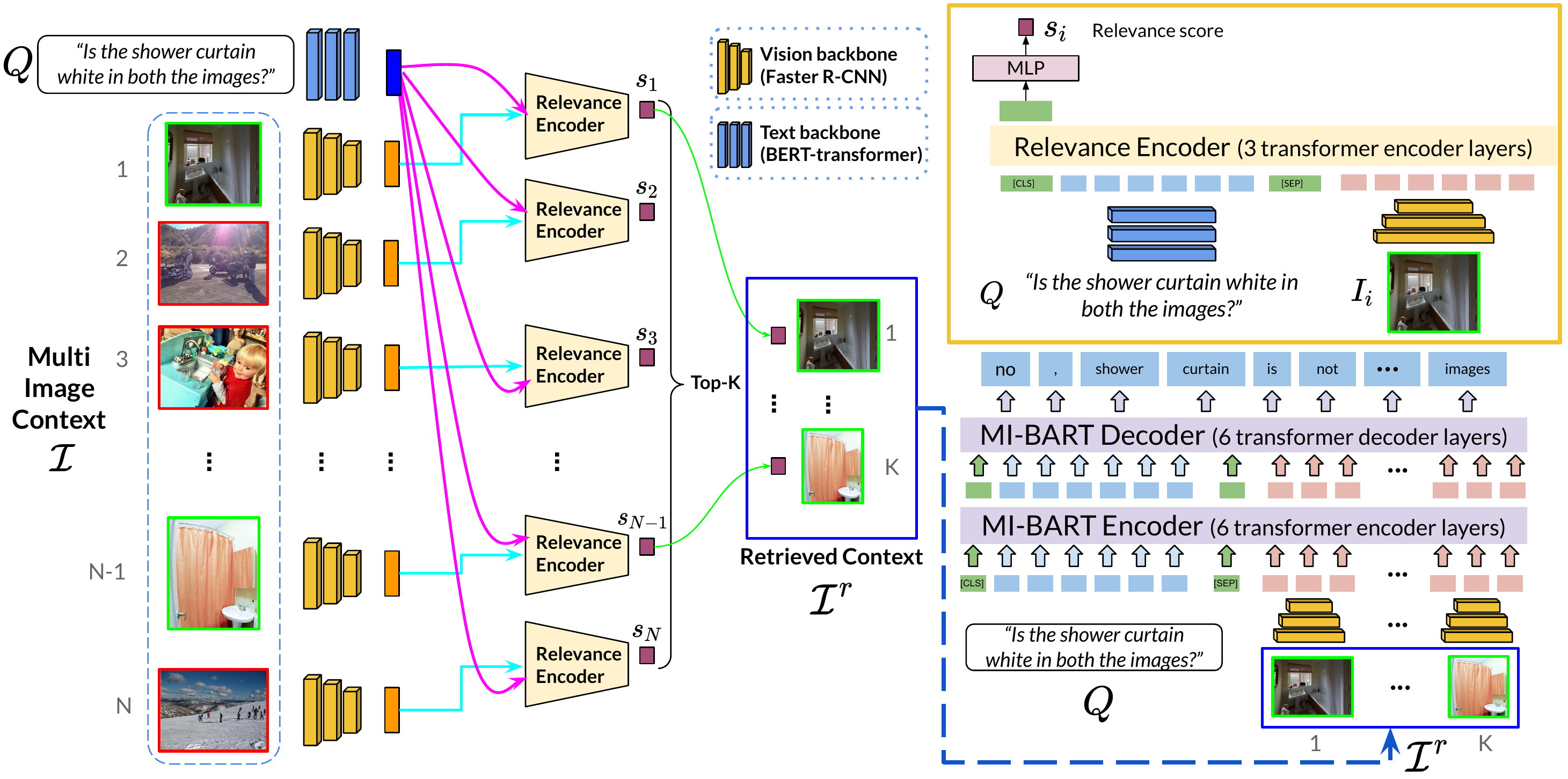}
\caption{{An overview of our proposed framework for retrieval-based VQA}. Given a question $Q$ and a pool of images $\mathcal{I}$, we encode the question and each image using a pretrained BERT and a pretrained Faster R-CNN, respectively. Once encoded, our multimodal relevance encoder (shown in the yellow box at the top right) generates relevance scores $S$ for all images in the set with the question. We choose top-$K$ scoring images as the retrieved relevant images $\mathcal{I}^r$. We encode the images in $\mathcal{I}^r$ using Faster R-CNN and feed them to our MI-BART encoder along with $Q$ to facilitate joint reasoning over the multi-image context with respect to the question. Once the MI-BART encoder encodes the question in the context of retrieved images, the MI-BART decoder generates the free-form natural language answer $A$ to the question. }
 \label{fig:main_arch}
\end{figure*}

To overcome such limitations, we curate a dataset \textsc{RetVQA} from Visual Genome, where we emphasize multi-image, metadata-independent questions over a pool of heterogeneous images, expecting a mix of classification-oriented and open-ended generative answers. 
We leverage question-answer and object annotations of Visual Genome to curate the dataset. We curate truly multi-image questions spanning over five different categories, namely, color, shape, count, object-attributes, and relation-based. For each question category, we curate binary-generative and open-ended generative answers. Dataset statistics are shown in Tables~\ref{tab:moredatasetStats} and~\ref{tab:datasetStats}.

The questions in \textsc{RetVQA} are curated as follows. We start by extracting subjects and relations of the existing question-answer pairs from Visual Genome; for example, consider these two question-answer pairs: $q_1$ (over image $I_1$): ``What is the cow eating in the image?'' where the answer is $a_1$: ``grass''; and, $q_2$ (over image $I_2$): ``what is the sheep eating?'' where the answer is $a_2$: ``grass''. Given $q_1$ and $q_2$, we extract subjects (``cow'' and ``sheep''), relations (``eating (eat/eats)''), and then we frame combined questions using templates (over images $I_1$ and $I_2$) as follows. $q_3$: ``what else eats the same thing as cow does?'' with answer $a_3$: ``sheep eats the same thing as cow''. Another question could be $q_4$: ``Does cow and sheep eat the same thing?'' where the answer is $a_4$: ``Yes, cow and sheep eat the same thing''. Thus, we curate binary-generative (like $q_4$) and open-ended generative (like $q_3$) types of answers. We further associate negative images for each of the curated questions using their object annotations as follows. A negative image is one where both the subject and object (used to generate the question) do not exist together in the image. This enforces that the answer has to be inferred only when all the relevant images are correctly retrieved and the negatives serve as sufficiently hard negatives.

We use a random 80\%-10\%-10\% train-validation-test split. All the questions in our dataset have at least two relevant images and 24.5 irrelevant images on average. A comparison with the other relevant datasets is shown in Table~\ref{tab:datasetComp}. Further, Figure~\ref{fig:datasetAnalysis} shows the distribution of unique answers across question types and question length distribution. Figure~\ref{fig:word_cloud} shows the word cloud of top-80 frequent answers. We observe that most questions are in the 5--10 words range, and there is no noticeable bias towards the majority of answers in the dataset.

\section{Retrieval QA Methodology}
\label{sec:webqa_method}

\subsection{\textsc{RetVQA} Problem Formulation} 
The \textsc{RetVQA} problem is defined as follows. Given a natural language question $Q$, a set of $N$ heterogeneous images $\mathcal{I} = \{I_1, I_2, \ldots, I_N\}$, the task is to generate an answer ($A$) for the question $Q$ based on $\mathcal{I}$ where only a few images are relevant for the question. To answer the question $Q$ using $\mathcal{I}$, we need a method that retrieves the relevant images $\mathcal{I}^r\subseteq \mathcal{I}$ and then leverages the retrieved context $\mathcal{I}^r$. Accordingly, our labelled dataset consists of quadruplets $(Q, \mathcal{I}, \mathcal{I}^r, A)$.

\subsection{\textsc{RetVQA} Framework}
The proposed framework solution: (i) multi-modal relevance encoder for retrieval of relevant sources $\mathcal{I}^r$ from $\mathcal{I}$ for the given question $Q$ and (ii) a unified \underline{M}ulti \underline{I}mage \underline{BART} (MI-BART) to generate fluent free-form natural language answer for the question $Q$ using the retrieved images $\mathcal{I}^r$ as context. 

\noindent\paragraph{Image representation.} Inspired by recent vision-language pretraining literature~\cite{visualbert,chen2020uniter,oscar}, for every image ${I_i}$ in $\mathcal{I}$ where $i\in\{1,2,\ldots, N\}$, we first detect a fixed set of $\mathcal{P}$ objects using Faster R-CNN~\cite{frcnn} pretrained on Visual Genome~\cite{krishna2017visual}. For every object $p$, where $p\in\{1,2,\ldots,\mathcal{P}\}$, we obtain $2048$-dimensional regional feature $\mathbf{o}_p^{reg}$ and $4$-dimensional bounding box co-ordinates $\mathbf{o}_p^{bbox}$. Thereby, each image $I_i$ 
is represented by a set of $\mathcal{P}$ object proposals $\{(\mathbf{o}^{reg},\mathbf{o}^{bbox})_p\}_i$. 
Following~\cite{visualbert}, for every region $p$ 
we project both $2048$-dimensional regional representation and $4$-dimensional bounding box coordinates into the $d$-dimensional space using a linear projection to obtain $\{\mathbf{o}_p\}_i$ and then concatenate across all regions within the image to obtain image embedding $\mathbf{o}_i$ as follows.
\begin{equation}
    \mathbf{o}_i = \{\mathbf{o}_p\}_i, \text{ where } p\in\{1,2,\ldots,P\} \text{, } i\in{1,2,\ldots,N}.
\end{equation}
\noindent\paragraph{Question representation.} We encode the textual question $Q$ containing $M$ words using a pretrained BERT~\cite{devlin2019bert}. This results into a sequence $\mathbf{q}$ of $M$ $d$-dimensional vectors, $\mathbf{q}=\{\mathbf{q_m}\}$ where $m\in\{1,2,\ldots, M\}$. Note that if any additional metadata is available (e.g. captions in WebQA dataset), we augment it to the question.

\begin{equation}
    \mathbf{q}=\{\mathbf{q_m}\}=BERT(Q), \text{where } m\in\{1,2,\ldots, M\}.
\end{equation}
\subsection{Multimodal Relevance Encoder for Image Retrieval}

\noindent\paragraph{Pretraining.}Our multi-modal Relevance Encoder (RE) consists of three transformer encoder layers followed by a multi-layered perceptron (MLP) with a sigmoid unit over the final representation of the $[CLS]$ token. We pretrain our relevance encoder on MS-COCO~\cite{lin2014microsoftCOCO} using two unsupervised objectives, Image Text Matching (ITM) and Masked Language Modelling (MLM) similar to~\cite{visualbert}.

\paragraph{Question-Image relevance learning.} Each sample in our dataset contains a question $Q$ and $N$ images $I_1, I_2, \ldots, I_N$ of which some have been labelled as positive and others negative. Further, for each image, we have $P$ regions. We use each question-image pair $(Q, I_i)$ to learn question-image relevance using our multi-modal Relevance Encoder (RE). Our pretrained multi-modal relevance encoder is fed with question-image pairs, along with two special tokens, $[CLS]$ and $[SEP]$; in short, the input to our relevance encoder is $[[CLS]$, $\{\mathbf{q_1}, \mathbf{q_2}, \ldots, \mathbf{q_m}\}$, $[SEP]$, $\{\mathbf{o_1}_i, \mathbf{o_2}_i, \ldots, \mathbf{o_p}_i\}$$]$. Our encoder then allows the input $M+P+2$ token sequence of question-image features to attend to each other and produces a sequence of contextualized embeddings. The $d$-dimensional contextualized embedding of $[CLS]$ token is further fed to an MLP with a sigmoid unit to produce a relevance score ($\hat{s}_i$) between 0 and 1 (Eq.~\ref{relevance_score}), indicating whether the given question-image pair $(Q, I_i)$ is relevant or not. We finetune our multi-modal relevance encoder parameters $\phi$ by minimizing binary cross-entropy loss $\mathcal{L}_{REL}(\phi)$ (Eq.~\ref{binary_loss_relevance}).
\begin{equation}
    \label{relevance_score}
    \hat{s}_i =  RE_\phi(Q, I_i).
\end{equation}
\begin{equation}
    \label{binary_loss_relevance}
    \mathcal{L}_{REL}(\phi) = -\E_{(Q, I_i) \sim D}[{s}_i\log({\hat{s}_i})+(1-{s}_i)\log(1-{\hat{s}}_i)].
\end{equation}

Given a question $Q$ and a set of $N$ images $\mathcal{I}$ sampled from our dataset $D$, we obtain relevance scores $S = \{{\hat{s}_i}\}_{i=1}^N$ for each question-image pair $(Q, \{I_i\}_{i=1}^N)$ using our fine-tuned relevance encoder (Eq.\ref{all_relevance_scores}). To choose the final set of relevant images $\mathcal{I}^r$ from the pool of images $\mathcal{I}$, we rank all the images in the pool using $S$ and choose top-$K$ images as our relevant context $\mathcal{I}^r$ for the given question $Q$ (Eq.~\ref{top_k_relevance_scores}).

\begin{equation}
    \label{all_relevance_scores}
    S = \{{\hat{s}_i}\}, \text{ where } {\hat{s}_i} = RE(Q, I_i), i\in{1,\ldots, N}.
\end{equation}
\begin{equation}
    \label{top_k_relevance_scores}
    \mathcal{I}^r = \{I_k\} \text{ where } k \in \text{top-}K(S).
\end{equation}

\subsection{\underline{M}ulti \underline{I}mage \underline{BART} for Question Answering}
Given the question and the retrieved images $\mathcal{I}^r$, the goal of MI-BART is to generate an accurate yet fluent free-form natural language answer for the question. Towards this end, we propose an encoder-decoder architecture similar to SimVLM~\cite{simvlm}. MI-BART encoder is a stack of six transformer layers~\cite{transformer_vaswani}, where each transformer layer comprises a self-attention layer, followed by a fully connected linear layer with a residual connection. Similarly, the MI-BART decoder is also a stack of six transformer layers~\cite{transformer_vaswani}, with an additional cross-attention layer in each transformer layer. We concatenate question embedding $\mathbf{q}$ with image embeddings of each image $I_k$ in $\mathcal{I}^r$ with a special token $[SEP]$ in between. Also, to distinguish the image features belonging to different images in the retrieved image set $\mathcal{I}^r$, we assign image order ids to image features from different images. Note that image order ids are not meant for assigning a sequence number to images in the retrieved set. Their sole purpose is to differentiate image features from different images. In short, the input to our MI-BART encoder is $[[CLS],$ $\{\mathbf{q_1}, \mathbf{q_2}, \ldots, \mathbf{q_m}\}$, $[SEP]$, $\{\mathbf{o_1}, \mathbf{o_2}, \ldots, \mathbf{o_p}\}_1$, $[SEP]$, \ldots, $\{\mathbf{o_1}, \mathbf{o_2}, \ldots, \mathbf{o_p}\}_k]$, where $m\in\{1,2,\ldots, M\}$, $p\in\{1,2,\ldots, P\}$ and $k\in\{1,2\ldots, K\}$. These inputs attend over each other through various self-attention layers of the MI-BART encoder and produce a sequence of contextualized embeddings $\mathbf{z} = \{\mathbf{z_j}\}$, where $j\in\{1,2,\ldots, M + (P \times k) + k\}$ (Eq.~\ref{mibart_encoder}).
\begin{equation}
    \label{mibart_encoder}
    \mathbf{z} = \{\mathbf{z_j}\} = \text{MI-BARTEncoder}(Q, \mathcal{I}^r).
\end{equation}

\begin{table*}[t]
  \centering
  \resizebox{0.7\textwidth}{!}{
  \begin{tabular}{l c c c c c c c c c c c c}
\hline
  \multicolumn{1}{c}{} & \multicolumn{6}{c}{\textsc{RetVQA}} &
  \multicolumn{6}{c}{WebQA} \\
  \cmidrule(r){2-7}
  \cmidrule(r){8-13}
   \multicolumn{1}{c}{} & \multicolumn{3}{c}{Oracle Images} &
   \multicolumn{3}{c}{Retrieved Images} & \multicolumn{3}{c}{Oracle Images} & \multicolumn{3}{c}{Retrieved Images} \\
   \cmidrule(r){2-4}
   \cmidrule(r){5-7}
   \cmidrule(r){8-10}
   \cmidrule(r){11-13}
  Method & Acc. & F & F$\times$A & Acc. & F & F$\times$A & Acc. & F & F$\times$A & Acc. & F & F$\times$A \\
\hline
    \textbf{Popularity-based Baselines} & & & & \\
   ~~~~Global popularity & 27.4 & 14.5 & 7.9 & 36.2 & 16.7 & 9.8 & 17.7 & 1.3 & 0.4 & 17.7 & 1.3 & 0.4 \\
   ~~~~Per-category popularity & 27.8 & 16.0 & 7.6 & 27.8 & 16.0 & 7.6 & 25.2 & 1.3 & 0.5 & 25.2 & 1.3 & 0.5\\
\hline
    \textbf{Other Baseline Approaches} & & & & \\
   ~~~~Question only & 62.4 & 15.3 & 10.4 & 62.4 & 15.3 & 10.4 & 22.2 & 34.9 & 13.4 & 22.2 & 34.9 & 22.2 \\
   ~~~~Aggregate VQA & 69.2 & 17.1 & 13 & 66.6 & 16.2 & 11.9 &*&*&*&*&*&*\\
   ~~~~VLP~\cite{VLP} & 65.1 & 70.2 & 58.8 & 65.1 & 70.2 & 58.8 & 45.7 & 42.2 & 25.9 & 44.2 & 38.9 & 24.1 \\
\hline
    \textbf{MI-BART (Ours)} & & & & \\
   ~~~~Image stitch \textbf{MI-BART} & 78.2 & 74.7 & 70.7 & 72.1 & 76.6 & 66.8 & 49.6 & 50.5 & 27.5 & \textbf{49.1} & 50.3 & 27.4 \\
   ~~~~\textbf{MI-BART} & \textbf{84.2} & \textbf{85.6}& \textbf{79.8} & \textbf{76.5} & \textbf{79.3} & \textbf{70.9} & \textbf{49.8} & \textbf{51.1} & \textbf{28.1} & 48.7 & \textbf{50.7} & \textbf{27.6} \\
\hline
\end{tabular}}
\caption{Performance comparison of various methods on \textsc{RetVQA} and image segment of WebQA. $^*$ WebQA only provides full-sentence answers rather than answer category annotations. Therefore, classification model like AggregateVQA cannot be trained for WebQA.}
\label{tab:mainRes-1}
\end{table*}

\begin{table*}[t]
    \centering
    \resizebox{0.8\textwidth}{!}{
    \begin{tabular}{l ccc ccc ccc ccc ccc }
\hline
    \multicolumn{1}{c}{} & \multicolumn{3}{c}{Color} & \multicolumn{3}{c}{Shape} & \multicolumn{3}{c}{Count} &
    \multicolumn{3}{c}{Object-attributes} & \multicolumn{3}{c}{Relation-based} \\ 
    \cmidrule(r){2-4}
    \cmidrule(r){5-7}
    \cmidrule(r){8-10}
    \cmidrule(r){11-13} 
    \cmidrule(r){14-16}
    Method & Acc. & F & F$\times$A & Acc. & F & F$\times$A & Acc. & F & F$\times$A & Acc. & F & F$\times$A & Acc. & F & F$\times$A \\
\hline
    \textbf{Popularity-based Baselines} & & & & \\
    ~~~~Global popularity & 25.4 & 8.2 & 0.6 & 24.5 & 9.2 & 1.3 & 25.6 & 13.4 & 6.7 & 49.5 & 35.2 & 30.5 & 0.0 & 4.8 & 0.0 \\
    ~~~~Per-category popularity & 25.3 & 9.1 & 0.5 & 24.5 & 9.2 & 1.3 & 24.5 & 14.4 & 4.5 & 49.5 & 35.2 & 30.5 & 6.0 & 15.6 & 2.3 \\ 
\hline
    \textbf{Other Baseline Approaches} & & & & \\
    ~~~~Question-only & 58.0 & 12.2 & 6.1 & 86.9 & 13.3 & 11.8 & 51.6 & 15.3 & 9.6 & 74.9 & 21.8 & 18.3 & 12.4 & 14.7 & 4.1 \\
    ~~~~Aggregate VQA & 60.1 & 12.4 & 6.7 & 91.3 & 14.4 & 13.5 & 54.6 & 15.6 & 10.3 & 75.4 & 21.9 & 18.6 & 32.2 & 20.8 & 11.9 \\
    ~~~~VLP~\cite{VLP} & 62.0 & 67.3 & 52.8 & 84.0 & 81.0 & 75.7 & 50.8 & 70.0 & 50.8 & 76.8 & 78.2 & 74.8 & 36.8 & 33.8 & 18.5 \\
 \hline
    \textbf{MI-BART (Ours)} & & & & \\
    ~~~~Image stitch \textbf{MI-BART} & 71.8 & \textbf{76.8} & 63.7 & \textbf{96.2} & \textbf{94.4} & \textbf{91.1} & 62.7 & \textbf{80.1} & 62.6 & \textbf{81.6} & \textbf{87} & \textbf{81.4} & 52.0 & 39.5 & 26.4 \\
    ~~~~\textbf{MI-BART} & \textbf{72.1} & 75.7 & \textbf{63.9} & 92.4 & 90.3 & 87.7 & \textbf{66.0} & 80.0 & \textbf{66.0} & 78.5 & 83.4 & 78.4 & \textbf{69.5} & \textbf{50.4} & \textbf{43.1}\\ 
\hline
  \end{tabular}}
    \caption{Performance breakdown for various methods by question categories on \textsc{RetVQA} with the retrieved images.}
  \label{tab:qa_resultsBreakDown}
\end{table*}

MI-BART decoder auto-regressively predicts the probability of the next token $A_t$ in the answer $A$ by attending to these encoder outputs $\mathbf{z}$ and previously generated answer tokens $a_{<t}$ through cross-attention and self-attention layers, respectively (Eq.~\ref{mibart_decoder}). We train MI-BART decoder parameters $\theta$ by minimizing the generative loss $\mathcal{L}_{GEN}(\theta)$ for generating the target answer token conditioned on the question $Q$ and retrieved image context $\mathcal{I}^r$ (Eq.~\ref{generative_loss}). During training, we leverage the ground truth relevant images as retrieved image context $\mathcal{I}^r$, while during inference, we obtain it from our relevance encoder.
\begin{equation}
    \label{mibart_decoder}
    P_\theta(A_t|A_{<t}, Q, \mathcal{I}^r) = \text{MI-BARTDecoder}(\mathbf{z}, A_{<t}).
\end{equation}
\begin{equation}
    \label{generative_loss}
    \mathcal{L}_{GEN}(\theta) = -\E_{(Q, \mathcal{I}^r)\sim D}\left[\sum_{t=1}^{|A|}\log(P_\theta(A_t|A_{<t}, Q, \mathcal{I}^r))\right].
\end{equation}

To summarize, our proposed framework works as follows, given a question $Q$ and a pool of $N$ images $\mathcal{I}$, we (i) obtain question-image relevance scores $S$ for each question-image $(Q, I_i)$ pair in $(Q, \mathcal{I})$ using our multimodal relevance encoder, (ii) rank all images in the pool based on $S$, and choose top-$K$ images as retrieved images context $\mathcal{I}^r$, and (iii) question $Q$ along with the retrieved image context $\mathcal{I}^r$ is fed to MI-BART which encodes the provided context and generates the free-form natural language answer $A$ to the question $Q$.

\subsection{Image-stitch MI-BART} 
Inspired by~\cite{imageset_vqa}, we consider stitching the retrieved images along the width into a single joint image and use single image VQA on this joint image. We use our proposed MI-BART for this baseline; however, we feed the stitched joint image instead of feeding $K$ images. While the MI-BART combines information across images in the embedding space, the image-stitch MI-BART combines information across images in the input space.

\section{Experiments and Results}
We conduct our experiments on \textsc{RetVQA} as well as on WebQA. Since our task deals with the image set as a given context, we consider the image-only subset of the WebQA dataset. Following~\cite{webQA21}, we use accuracy ($A$), fluency ($F$), and $F \times A$, as metrics to evaluate the generated answers. Accuracy validates whether the correct answer is present in the generated answer, whereas fluency measures the quality of the answer paraphrase. Fluency is computed using a recently proposed natural language generation metric called BARTScore~\cite{bartscore}. 
Further, an F1 score is used for retrieving relevant images from a pool of images.

\begin{figure*}[!t]
  \includegraphics[width=0.97\textwidth]{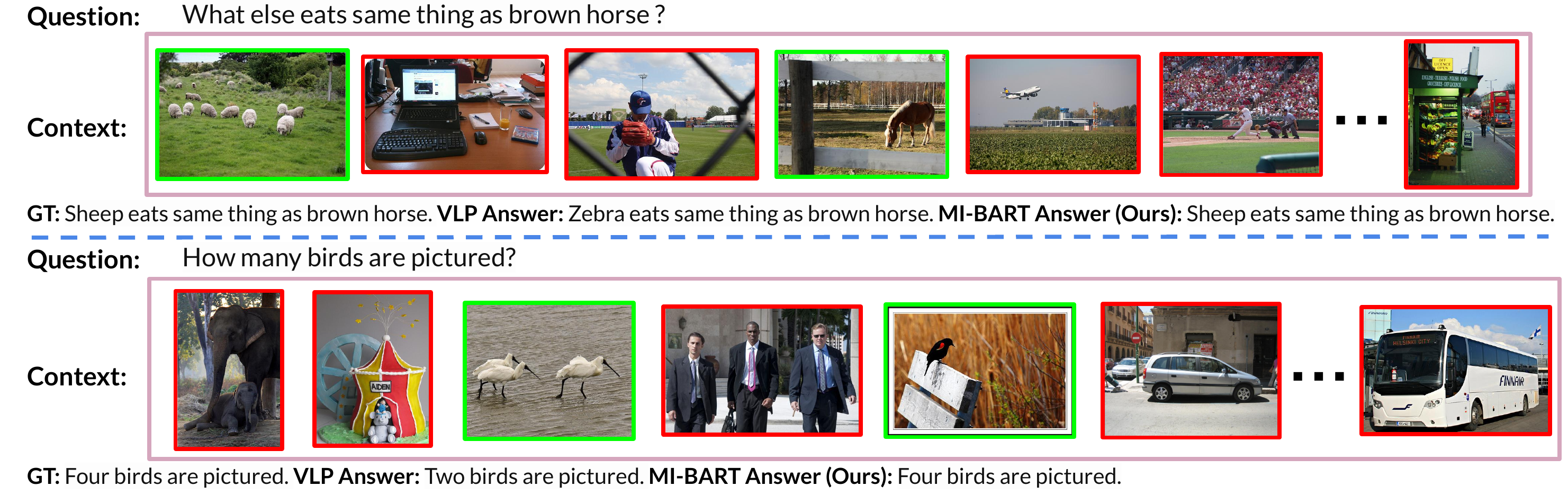}
  \caption{A selection of QA using MI-BART (Ours) and VLP (best baseline) from \textsc{RetVQA} test dataset. The images in the green and red bounding boxes show relevant and irrelevant images, respectively. Our approach consistently performs better than the baseline on different types of questions requiring multiple images to arrive at an accurate answer. (Best viewed in color).}
  \label{fig:visual_examples}
\end{figure*}

\begin{table}[t!]
    \centering
    \resizebox{0.9\columnwidth}{!}{
    \begin{tabular}{l ccc ccc}
    \hline
    \multicolumn{1}{c}{} & \multicolumn{3}{c}{Binary} & \multicolumn{3}{c}{Open-ended} \\ 
    \cmidrule(r){2-4}
    \cmidrule(r){5-7}
    Method & Acc. & F & F$\times$A & Acc. & F & F$\times$A \\
    \hline
    \textbf{Popularity-based Baselines} & & & & \\
    ~Global popularity & 49.9 & 17.3 & 14.3 & 0.0 & 11.1 & 0.0 \\
    ~Per-category popularity & 49.5 & 17.1 & 13.3 & 1.2 & 14.6 & 0.5 \\ 
    \hline
    \textbf{Other Baseline Approaches} & & & & \\
    ~Question-only & 74.2 & 11.5 & 9.2 & 48 & 20 & 12\\
    ~Aggregate VQA & 75.6 & 11.5 & 9.5 & 55.5 & 22 & 14.9 \\ 
    ~VLP~\cite{VLP} & 73.2 & 80 & 72.5 & 55.1 & 58.2 & 42 \\
    \hline
    \textbf{MI-BART (Ours)} & & & & \\
    ~Image stitch \textbf{MI-BART} & \textbf{80.4} & \textbf{88.3} & \textbf{80.3} & 69.4 & 70.2 & 57 \\ 
    ~\textbf{MI-BART} & 78.7 & 85.6 & 78.7 & \textbf{73.7} & \textbf{71.7} & \textbf{61.5} \\ 
    \hline
  \end{tabular}}
    \caption{Performance breakdown by answer categories for various methods on \textsc{RetVQA} with the retrieved images.}
  \label{tab:qa_resultsAnswerCategories}
\end{table}

\subsection{Baselines, Ablations and Implementation Details}
\paragraph{Baselines.} We compare our proposed method (MI-BART) and its variant image-stitch MI-BART with the following baselines: 
\noindent\textbf{(i) Popularity-based Baselines}: To check for prior biases associated with frequent answers globally or per question category, we use two popularity-based baselines. (a) Global popularity: the most frequent answer in the training set is always considered as the answer by the model, and (b) Per-category popularity: the most frequent answer for each question category is always considered the answer for questions in the corresponding question category. 
\noindent\textbf{(ii) Aggregate VQA}: \textsc{RetVQA} task involves VQA over multiple images. In the Aggregate VQA baseline, we use the traditional single-image VQA method~\cite{antol2015vqa} for each image and aggregate the results. Given a question $Q$ and its corresponding $K$ retrieved images, $\mathcal{I}^r$ from our relevance encoder, we feed each retrieved image along with the question $Q$ to a single image VQA model to get a joint representation. We concatenate joint representations of all retrieved images into a single representation $F$ and feed to a linear layer (MLP) to predict the final answer, i.e., $A = MLP(F)$. Since traditional VQA methods follow a classification-style answer prediction approach, we use the 1000 most frequent answers as classes in the softmax layer. To generate a fluent answer, we prepend the predicted answer to the question after removing the first word from the question. This baseline is not benchmarked on the WebQA dataset, as the dataset does not provide precise answer annotations for the trainset questions. \noindent\textbf{(iii) VLP}: As the \textsc{RetVQA} task requires the model to generate the text, encoder-only multimodal transformer models like ViLBERT~\cite{vilbert}, VisualBERT~\cite{visualbert}, OSCAR~\cite{oscar}, ViLT~\cite{vilt} and UNITER~\cite{chen2020uniter} are not directly suitable. Hence, we use VLP~\cite{VLP} which is a unified encoder-decoder multimodal transformer as our baseline. We finetune a pretrained VLP on our datasets for evaluation.

\begin{table}[!t]
\centering
    \resizebox{0.6\columnwidth}{!}{
    \begin{tabular}{l ccc}
    \hline
    Retrieval & Acc. & F & F$\times$A \\
    \hline
    Top-1 retrieved image & 59.8 & 61.1 & 48.8 \\ 
    All retrieved images & \textbf{76.5} & \textbf{79.3} & \textbf{70.9} \\
    \hline
  \end{tabular}}
    \caption{MI-BART performance on \textsc{RetVQA} using different retrieval strategies.}
  \label{tab:qa_resultsretrieval_ablation}

\end{table}

\paragraph{Ablations.} We perform the following ablations to better understand the various components of our proposed model. (i) \textit{Question-only}: To study the role of the images in generating accurate and fluent answers, we ignore the images and use questions only as input to our model. (ii) \textit{Single-image retrieval}: To study the importance of reasoning over multiple images to generate an answer to the question, we use top-1 retrieved image as our only context instead of a multi-image context. (iii) \textit{Missing captions}: To study the role of image metadata in the relevant source image retrieval and, thereby, the answer generation, we conduct experiments on WebQA without leveraging the image metadata (captions). In this ablation, we augment captions (available in WebQA) as part of the textual input in both the relevance encoder and MI-BART. 

\begin{table}[!t]
\centering
    \resizebox{0.57\columnwidth}{!}{
    \begin{tabular}{l c c c c}
    \hline
    \multicolumn{1}{c}{} & \multicolumn{3}{c}{Retrieval} & \multicolumn{1}{c}{\textbf{QA}} \\ 
    \cmidrule(r){2-4}
    \cmidrule(r){5-5}
    Captions & P & R & F1 & F$\times$A\\
    \hline
    w/ captions & 32.3 & 44.7 & 37.5 & 19.7\\ 
    w/o captions & \textbf{79.7} & \textbf{86.3} & \textbf{77.4} & \textbf{28.1} \\ 
    \hline
  \end{tabular}}
    \caption{Effect of w/ and w/o captions in WebQA: Performance of MI-BART on retrieval and QA (retrieved images setting).}
  \label{tab:webqaCaptionAblation}

\end{table}

\noindent\paragraph{Implementation details.} We have implemented our framework in PyTorch~\cite{paszke2019pytorch} and Hugging Face's transformers~\cite{wolf-etal-2020-transformers} library. Our relevance encoder has three transformer layers, each having eight attention heads. We pretrain our relevance encoder on MS-COCO~\cite{lin2014microsoftCOCO} with a constant learning rate of 1e-4 using Adam optimizer~\cite{kingma2014adam}. Using the same optimiser, we finetune the relevance encoder on both datasets with a constant learning rate of 2e-5. Our MI-BART has six standard transformer encoder layers and six standard transformer decoder layers~\cite{transformer_vaswani}; we initialize our MI-BART with VLBart~\cite{vlbart} pretrained weights to leverage the strong visual-textual learning of VLBart. We further finetune MI-BART on a multi-image QA task with a learning rate of 5e-5 using Adam optimizer with a linear warm-up of 10\% of the total steps. Our relevance encoder and MI-BART were trained using 3 Nvidia RTX A6000 GPUs with a batch size of 96 and 256 while training and a batch size of 360 and 480 during testing, respectively.

\begin{figure*}[t!]
\scriptsize
   \centering
  \includegraphics[width=0.97\textwidth]{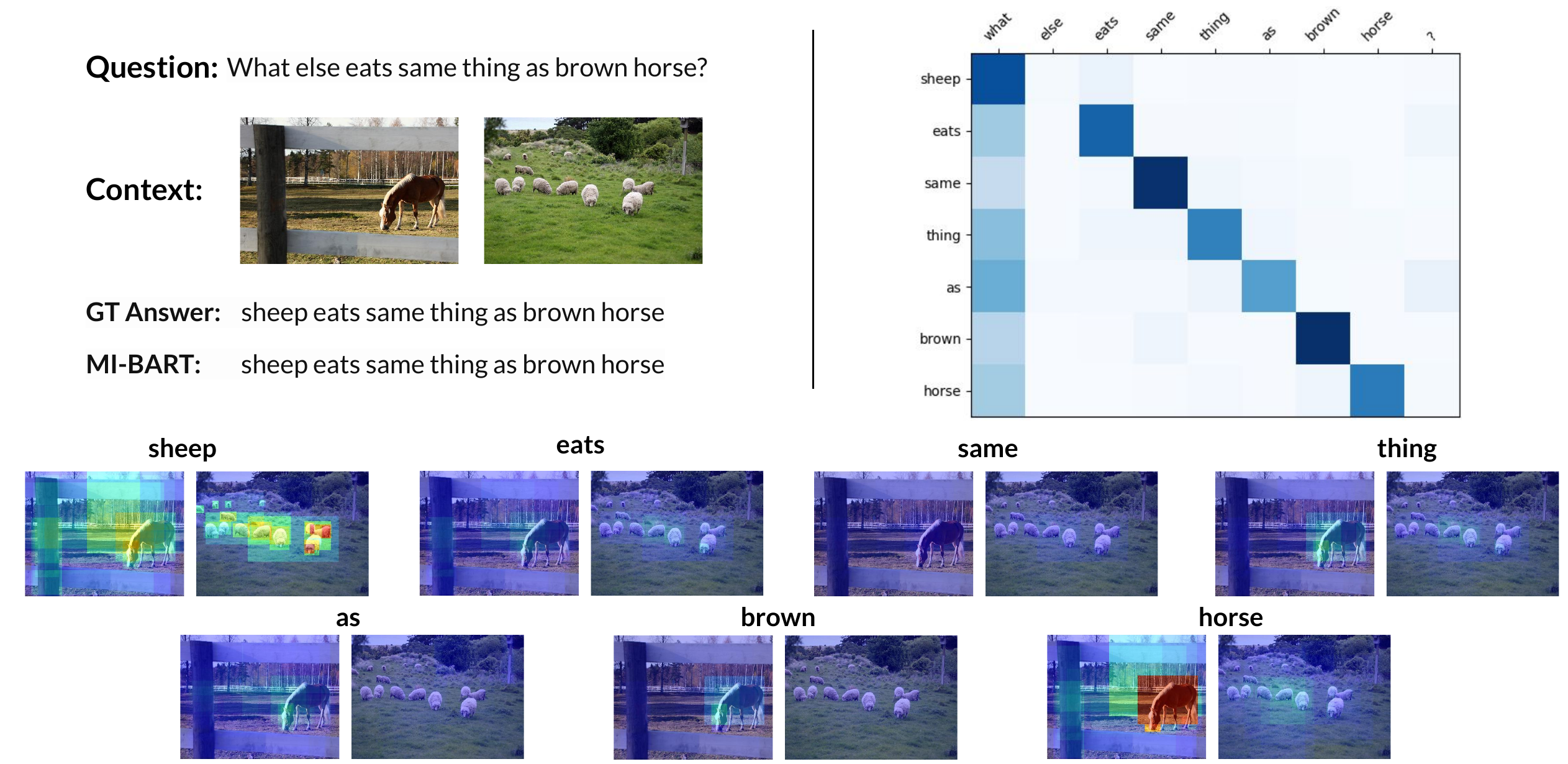}
\caption{Multimodal attention map over the retrieved images and the question during the answer generation. We observe that our proposed MI-BART attends to the relevant regions of both images while generating the main answer word `sheep'. Further, we see that it attends to brown horse regions of the first image along with the corresponding question parts while generating `brown horse'. [Special tokens are removed for visualization.] {(Best viewed in color).}}
 \label{fig:attention_map}
\end{figure*}

\subsection{Results and Analysis} 
We conduct our experiments in two settings, namely, (i) Oracle images: Here, we use ground-truth relevant images for answer generation, and (ii) Retrieved images: Here, relevant images are retrieved using our relevance encoder. We show the results under both these settings in Table~\ref{tab:mainRes-1}. We observe that the popularity-based methods perform poorly. This result is expected as popularity-based methods do not use any question or image context. Methods that involve either questions or images perform better than the popularity-based baselines. However, the question-only baseline has a F$\times$A score of 10.4 on \textsc{RetVQA}, showing that image context is needed to generate accurate yet fluent answers. Transformer-based baseline VLP and image-stitch MI-BART reach a F$\times$A score of 58.8 and 70.7 on our dataset, respectively, in the oracle setting, compared to 79.8 of our proposed MI-BART framework. Image-stitch MI-BART outperforms transformer-based VLP by 12\% on our dataset and 1.5\% on the WebQA dataset, which shows that having a separate decoder in the proposed MI-BART baseline has better reasoning capabilities than a unified encoder-decoder like VLP. We further present the QA results over the retrieved images setting using our relevance encoder, which has an F1 score of 71 at the top-2. All the approaches involving image context outperform question-only baseline, emphasizing that \textsc{RetVQA} has a reasonable utility to develop and benchmark methods capable of jointly reasoning over multi-image context and the question.

Further, in Table~\ref{tab:qa_resultsBreakDown}, we show the QA results over various question categories, and in Table~\ref{tab:qa_resultsAnswerCategories}, we show the results over answer categories under the retrieved images setting. Our framework outperforms baselines, especially in questions with open-ended generative answers, which constitute nearly half of our dataset. As expected, open-ended generative questions are more challenging than binary ones. However, compared to the baselines, MI-BART provides better improvement for open-ended questions than binary ones by jointly reasoning over multi-image context. Results in Table~\ref{tab:qa_resultsretrieval_ablation} further emphasize our hypothesis of requiring multiple images to answer the given question. We show the missing caption ablation results on the image-subset of WebQA in Table~\ref{tab:webqaCaptionAblation}; this result further affirms our claims that the performance of methods on the WebQA dataset depends on the image metadata like captions.

\noindent\paragraph{Qualitative analysis.}
We illustrate a selection of results using our proposed approach and one of the most competitive baselines viz. VLP in Figure~\ref{fig:visual_examples}. In both these results, our approach correctly answers the question in a large heterogeneous visual context. Further, to understand the importance of the multimodal input for question answering, we plot the multimodal attention map over the retrieved images and the question during the answer generation in Figure~\ref{fig:attention_map}. The figure shows that our proposed MI-BART model attends to the relevant regions of both images while generating the main answer word `sheep'. Further, we observe that it attends to brown horse regions of the first image along with the corresponding question parts while generating `brown horse'. Thus, both images are needed and paid attention to when generating the right answer. We further conducted a detailed error analysis on 50 randomly chosen samples where our model failed to generate a correct answer. We categorize the errors into four major categories: (i) partial retrieval: images retrieved by relevance encoder are partially relevant (52\%). (ii) Incorrect retrieval: images retrieved by the relevance encoder are entirely irrelevant (26\%). (iii) Incorrect reasoning: model generating a partially incorrect answer despite all the retrieved images being relevant (40\%). 


\section{Conclusion and Future Scope}
In this paper, we introduced the \textsc{RetVQA} task. We proposed a unified Multi Image BART model to answer the question from the retrieved images using our relevance encoder. Our proposed framework shows promising improvements over the baselines. We have also performed several ablations to further understand the importance of various modules in the proposed framework. In the future, we would like to explore stronger retrieval models and QA on a large pool of images. We firmly believe \textsc{RetVQA} will pave the way for further research avenues in a broader theme of web image QA.

\section*{Acknowledgments}
Abhirama is supported by the Prime Minister Research Fellowship (PMRF), Government of India. We thank Microsoft for supporting this work through the Microsoft Academic Partnership Grant (MAPG) 2021.

\small
\bibliographystyle{named}
\bibliography{ijcai23}

\clearpage
\appendix
\paragraph{\huge{Appendix}}
\section{Varying irrelevant images in \textsc{RetVQA} experiment}
We evaluate the proposed framework MI-BART on \textsc{RetVQA} with the varying numbers of irrelevant images in the pool. We report F1 on the retrieval task, along with accuracy, fluency and F $\times$ A metrics for question answering in Table~\ref{tab:qa_results_vary_negs}. 
\begin{table}[h!]
    \centering
    \scriptsize
    \begin{tabular}{l c|ccc }
    \hline
    \multicolumn{1}{c}{} & {\textbf{Retrieval}} & \multicolumn{3}{c}{\textbf{QA}}  \\  
    \cline{2-2}
    \cline{3-5}
    {\#Irrelevant images} & F1 & Acc. & F & F$\times$A \\
    \hline
    25 & 71.0 & 76.5 & 79.3 & 70.9 \\
    50 & 60.3 & 73.3 & 76.9 & 67.5\\
    100 & 48.4 & 69.8 & 74.2 & 63.5 \\
    200 & 36.3 & 66.9 & 71.8 & 60.2\\
    \hline
  \end{tabular}
    \caption{MI-BART performance on \textsc{RetVQA} with the varying number of irrelevant images in the pool.}
  \label{tab:qa_results_vary_negs}
\end{table}

As expected, we observe that retrieval F1 drops as we increase the pool size. However, thanks to the robustness of our proposed question answering approach, MI-BART, F$\times$A does not drop by the same extent. 

\section{Results on paraphrased questions}
\begin{figure}[h!]
\centering
  \includegraphics[width=\columnwidth]{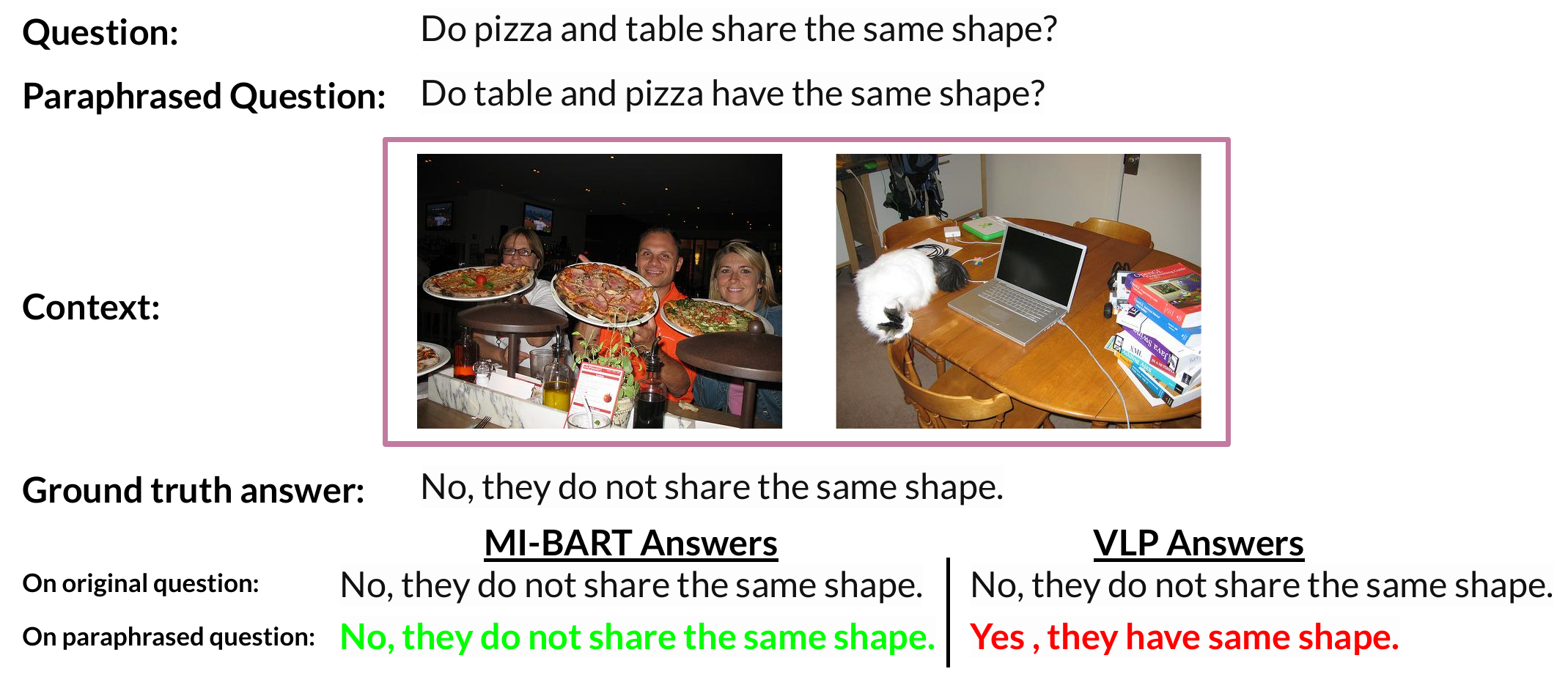}
  \caption{Paraphrased question answering using MI-BART (Ours) and VLP from \textsc{RetVQA} dataset.}
  \label{fig:paraphrase_example_figure}
\end{figure}

We wished to check if the high accuracies obtained by the proposed model is an outcome of the limited encoding of question templates in our \textsc{RetVQA} dataset. Hence, we evaluate our proposed framework (MI-BART) on 2K paraphrased questions in the test set with our strongest baseline (VLP). A subset of test set questions is paraphrased using the BART~\cite{bart} paraphrase model (Large)\footnote{\url{https://huggingface.co/eugenesiow/bart-paraphrase}}. We show the results in Table~\ref{tab:paraphrase_results}. Our results show that our proposed framework MI-BART is robust to the question templates used to create the data and performs fairly well on the paraphrased questions when compared against another transformer-based baseline VLP. We show a qualitative sample in Figure.~\ref{fig:paraphrase_example_figure}, where our model still gives the correct answer to the paraphrased question, whereas VLP entirely generates a wrong answer.

\section{Word Clouds for \textsc{RetVQA}}
\begin{figure*}[t!]
\begin{minipage}{0.33\textwidth}
    \centering
    \includegraphics[width=\columnwidth, height=\columnwidth]{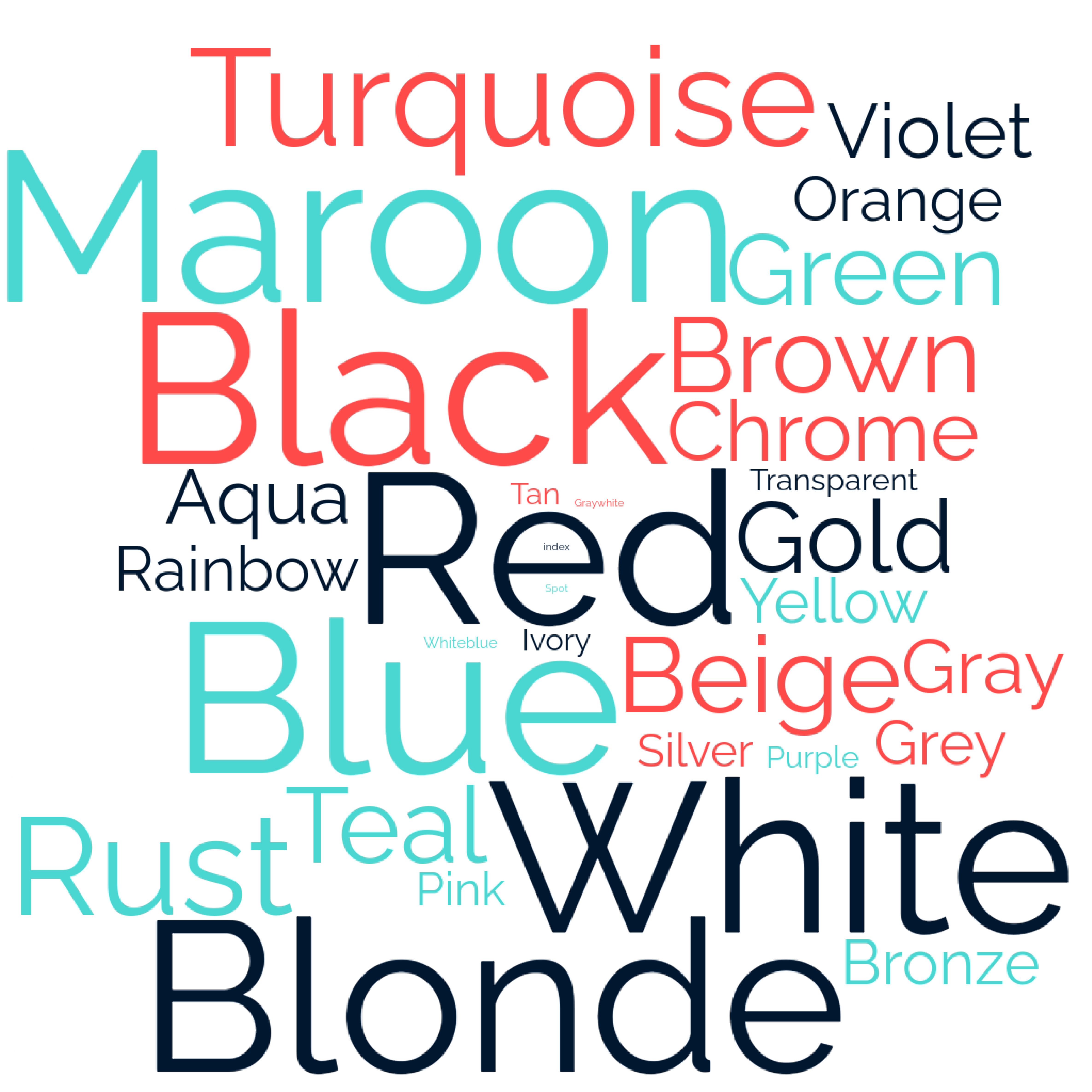}
\end{minipage}
\hspace{1pt}
    \begin{minipage}{0.33\textwidth}
    \centering
        \includegraphics[width=\columnwidth, height=\columnwidth]{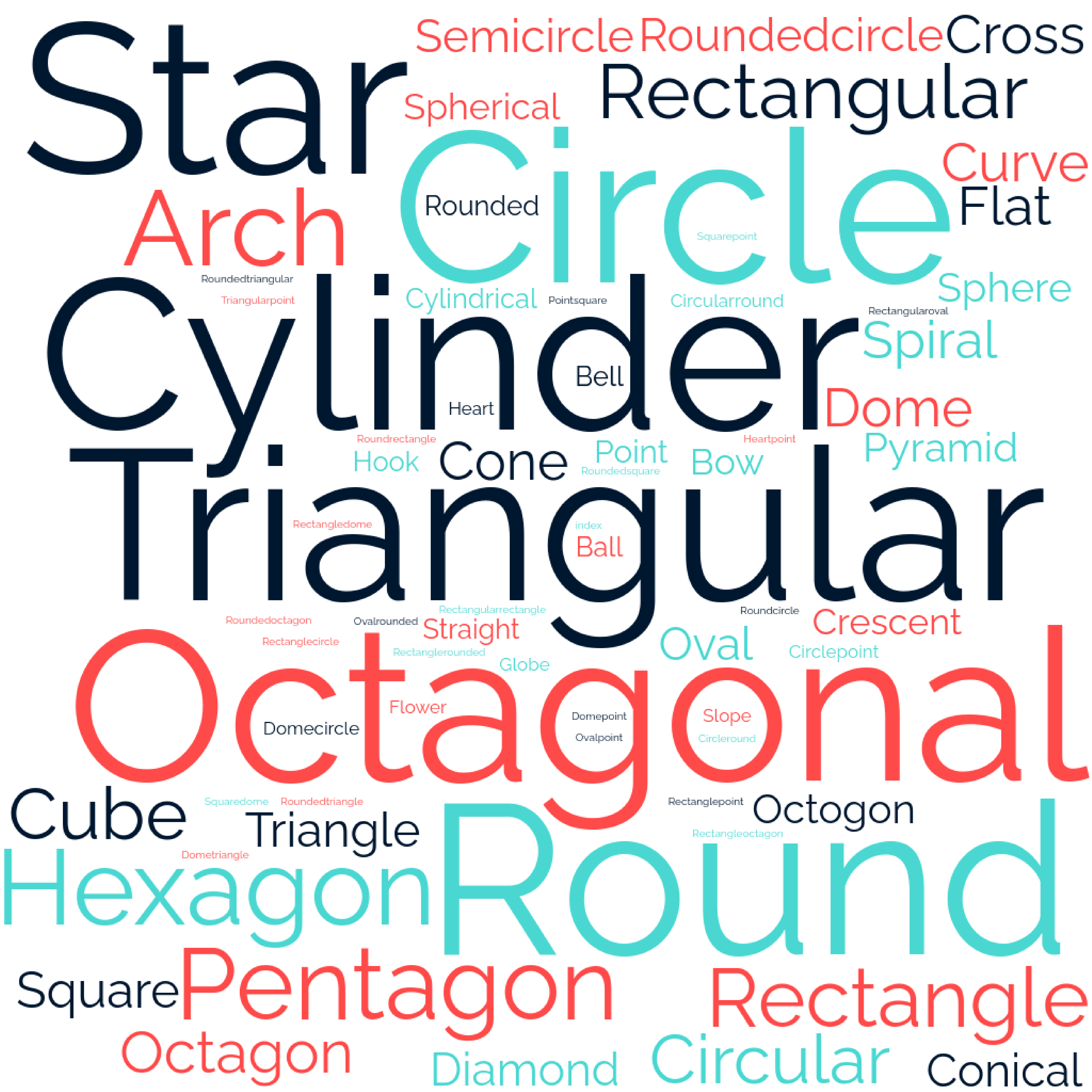}
    \end{minipage}
    \hspace{1pt}
     \begin{minipage}{0.33\textwidth}
    \centering
        \includegraphics[width=\columnwidth, height=\columnwidth]{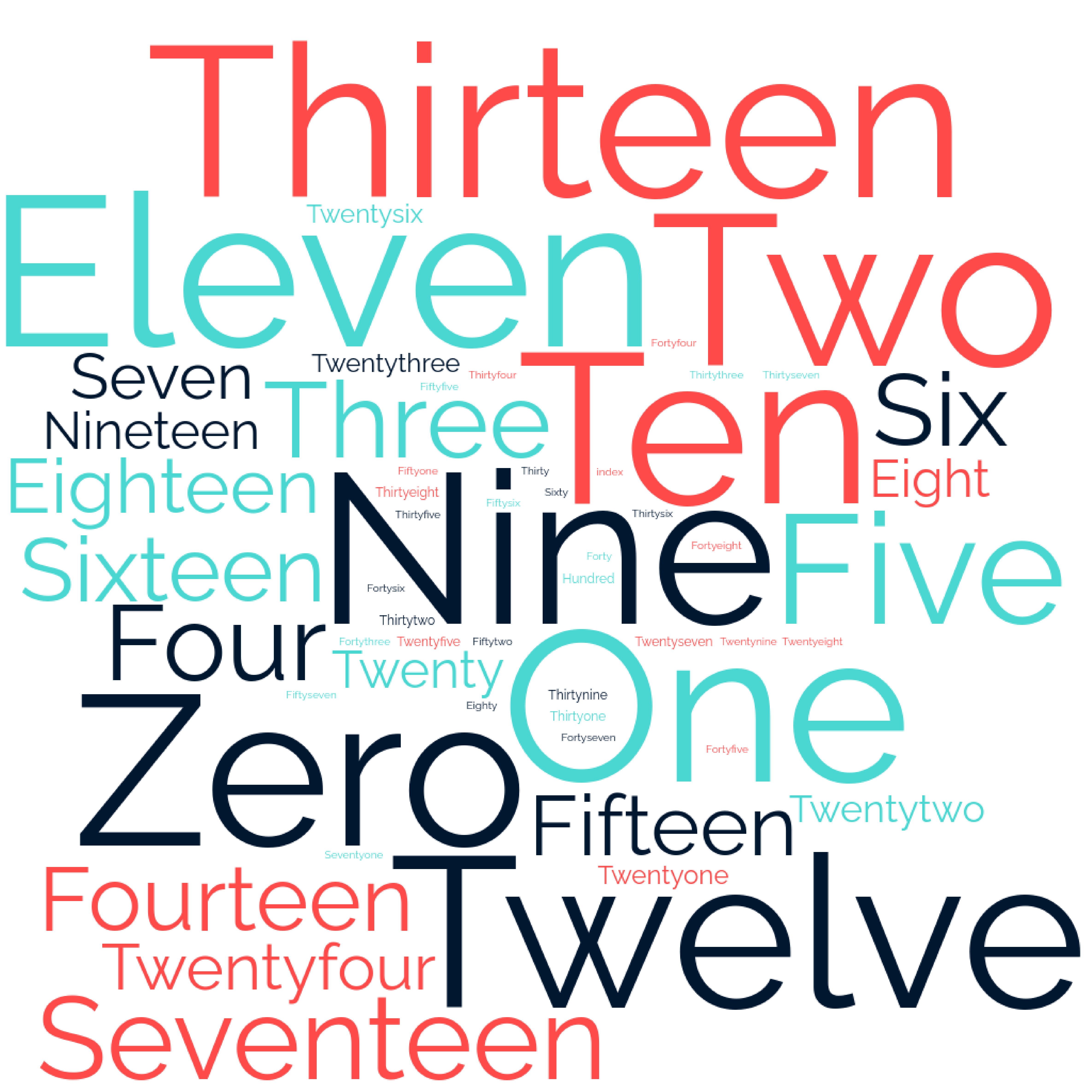}
    \end{minipage}
    \hspace{1pt}
        \caption{Word cloud for color, shape, count type question categories respectively (left to right) in the \textsc{RetVQA} dataset.}
    \label{fig:wordCloud}
\end{figure*}

Figure~\ref{fig:wordCloud} shows word clouds for popular short answers for the color, shape and count question categories, respectively (left to right) in the \textsc{RetVQA} dataset. There is a large coverage across different unique values showing that the dataset does not suffer from any majority bias across various question categories.

\begin{table}[h!]
    \centering
    \scriptsize
    \begin{tabular}{l ccc|ccc}
    \hline
    \multicolumn{1}{c}{} & \multicolumn{3}{c}{\textbf{VLP}} & \multicolumn{3}{c}{\textbf{MI-BART}}  \\  
    \cline{2-4}
    \cline{5-7}
    Question-type & Acc. & F & F$\times$A & Acc. & F & F$\times$A  \\
    \hline
    Paraphrased & 47.8 & 54.5 & 37.5 & \textbf{58.8} & \textbf{71} & \textbf{51.2} \\ 
    Non-paraphrased & 60.9 & 65.6 & 52.7 & \textbf{76.4} & \textbf{75.4} & \textbf{67.4} \\
    \hline
  \end{tabular}
    \caption{Performance on \textsc{RetVQA} with paraphrased questions with the retrieved images.}
  \label{tab:paraphrase_results}
\end{table}

\begin{figure*}
\centering
\begin{subfigure}[b]{\textwidth}
   \includegraphics[width=1\textwidth]{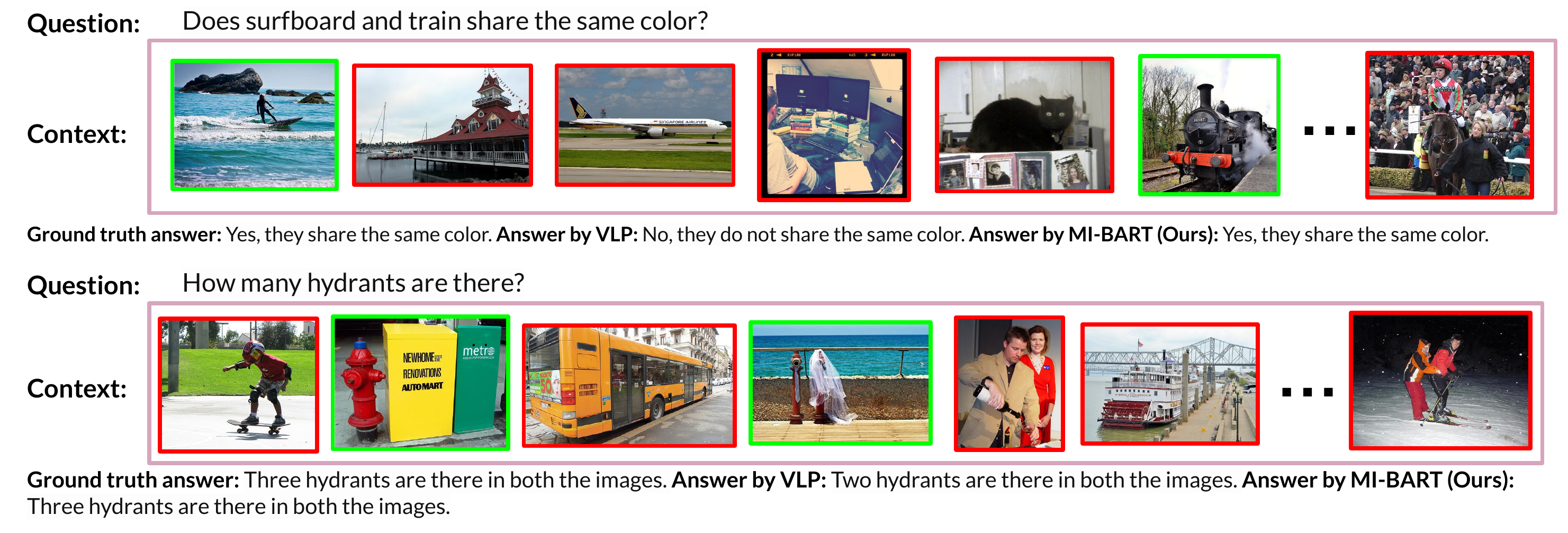}
\end{subfigure}
\begin{subfigure}[b]{\textwidth}
   \includegraphics[width=1\textwidth]{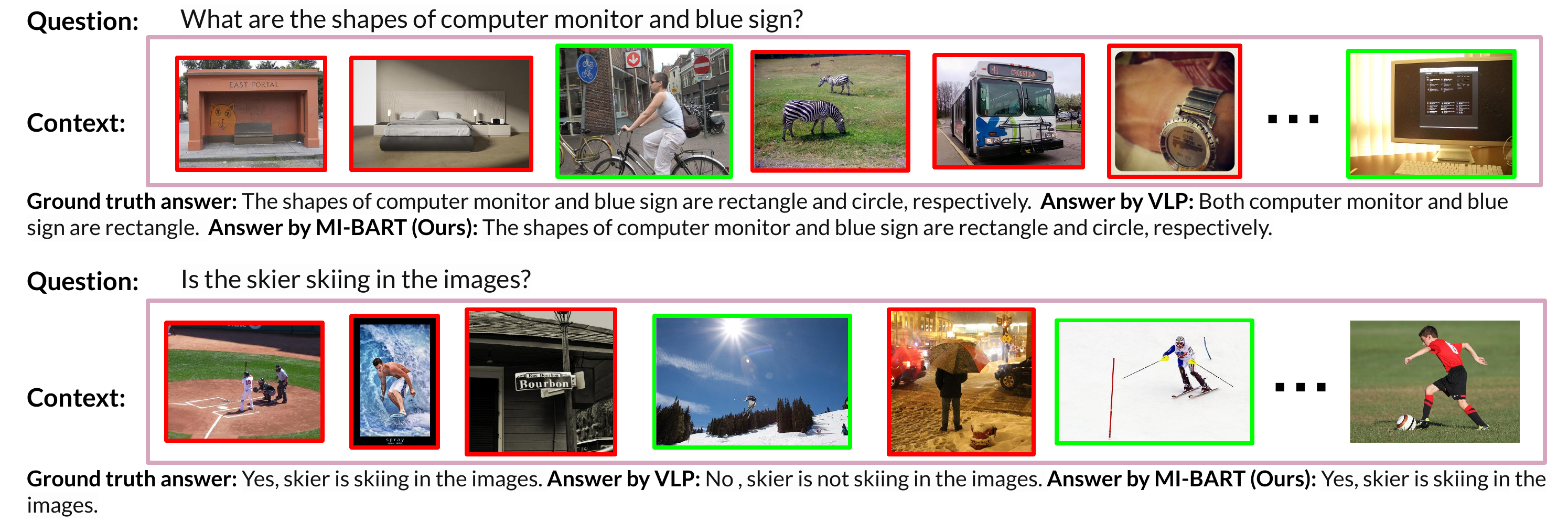}
\end{subfigure}
\caption{Few more selected example predictions using MI-BART (ours) and VLP (best baseline) from \textsc{RetVQA} dataset.}
\label{fig:Appendix_ex1}
\end{figure*}

\section{More examples}
\begin{figure*}
\centering
\begin{subfigure}[b]{\textwidth}
   \includegraphics[width=1\textwidth]{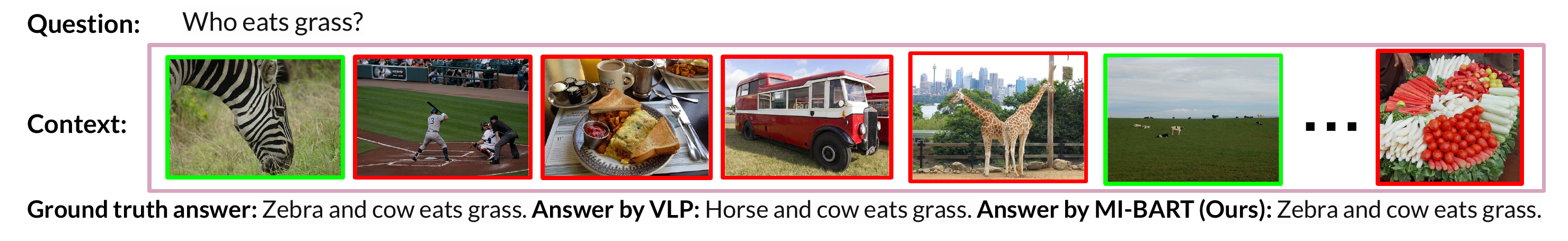}
\end{subfigure}
\begin{subfigure}[b]{\textwidth}
   \includegraphics[width=1\textwidth]{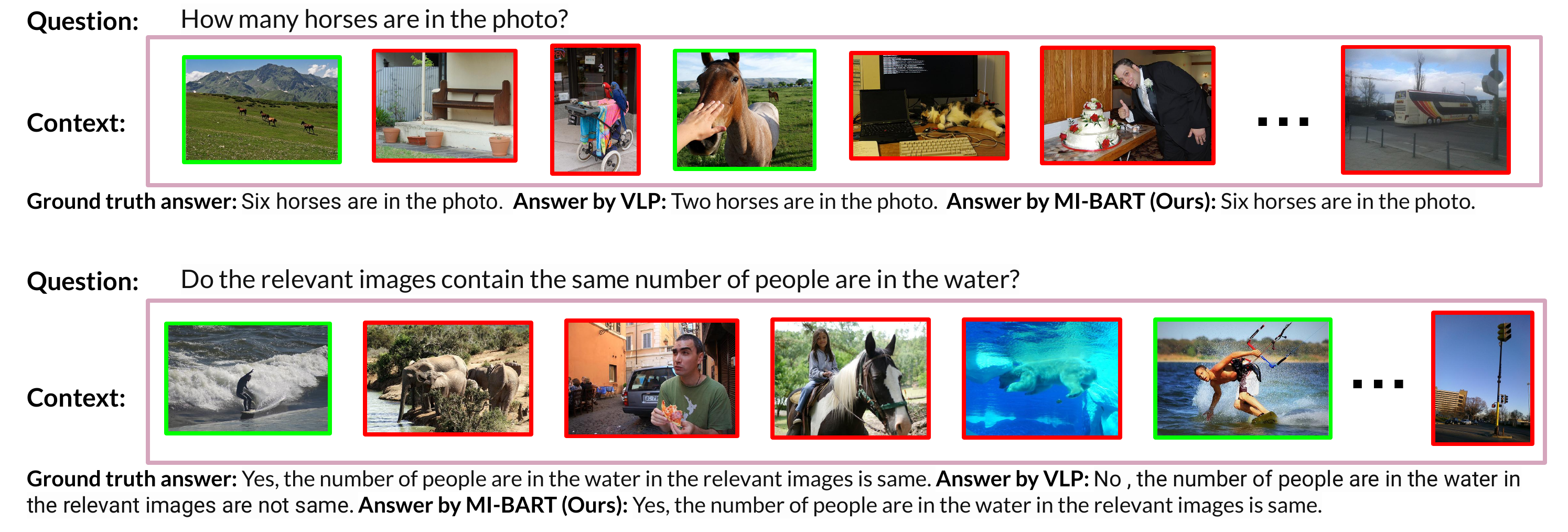}
\end{subfigure}
\begin{subfigure}[b]{\textwidth}
   \includegraphics[width=1\textwidth]{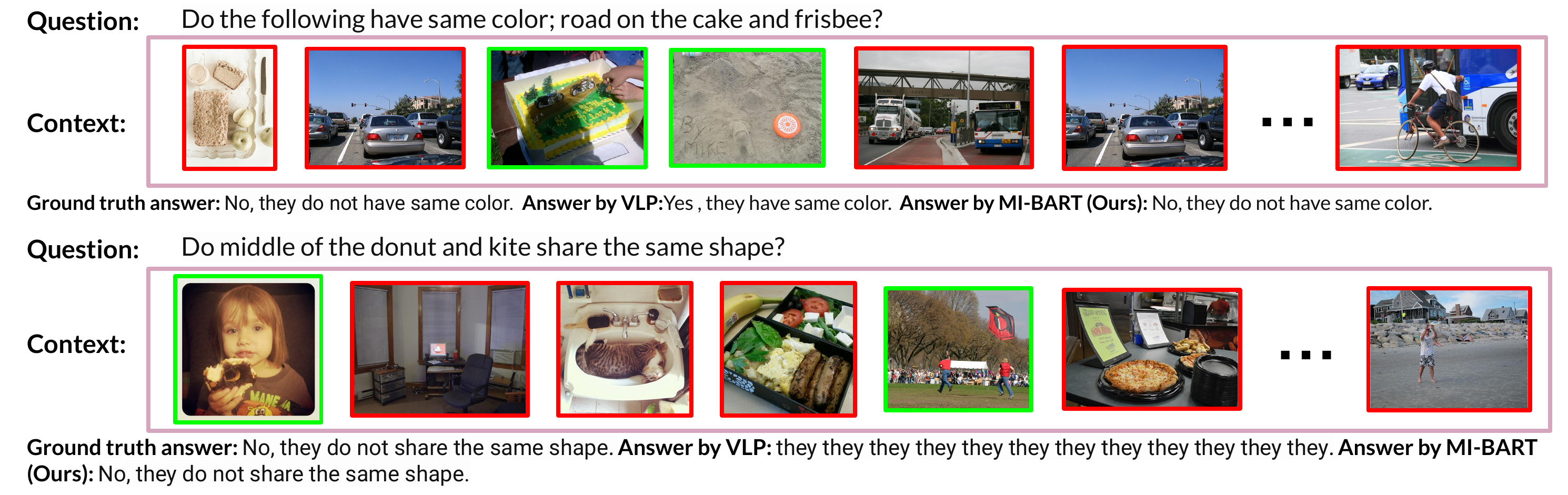}
\end{subfigure}
\caption{Few more selected example predictions using MI-BART (ours) and VLP (best baseline) from \textsc{RetVQA} dataset.}
   \label{fig:Appendix_ex2} 
\end{figure*}

We show some examples of predictions using MI-BART (ours) and VLP (best baseline) from \textsc{RetVQA} dataset in Figure~\ref{fig:Appendix_ex1} and Figure~\ref{fig:Appendix_ex2}.

\end{document}